\documentclass[letterpaper]{article} 
\usepackage{aaai24}  
\usepackage{times}  
\usepackage{helvet}  
\usepackage{courier}  
\usepackage[hyphens]{url}  
\usepackage{graphicx} 
\usepackage{natbib}  
\usepackage{caption} 
\usepackage{algorithm}
\usepackage{algorithmic}
\usepackage{graphicx}
\usepackage{amsmath}
\usepackage{amssymb}
\usepackage{booktabs}
\usepackage{multibib}
\usepackage{caption}
\usepackage{subfig}
\usepackage{multirow}

\usepackage[table]{xcolor}
\usepackage[british, english, american]{babel}

\newcites{app}{Appendix References}

\newlength\savewidth\newcommand\shline{\noalign{\global\savewidth\arrayrulewidth
  \global\arrayrulewidth 1pt}\hline\noalign{\global\arrayrulewidth\savewidth}}
\newcommand{\tablestyle}[2]{\setlength{\tabcolsep}{#1}\renewcommand{\arraystretch}{#2}\centering\footnotesize}

\makeatletter\renewcommand\paragraph{\@startsection{paragraph}{4}{\z@}
  {.5em \@plus1ex \@minus.2ex}{-.5em}{\normalfont\normalsize\bfseries}}\makeatother

\newcolumntype{x}[1]{>{\centering\arraybackslash}p{#1pt}}
\newcolumntype{y}[1]{>{\raggedright\arraybackslash}p{#1pt}}
\newcolumntype{z}[1]{>{\raggedleft\arraybackslash}p{#1pt}}

\newcommand{\app}{\raise.17ex\hbox{$\scriptstyle\sim$}}

\definecolor{baselinecolor}{gray}{.9}

\newcolumntype{*}{>{\global\let\currentrowstyle\relax}}
\newcolumntype{^}{>{\currentrowstyle}}

\definecolor{dt}{gray}{0.7}  %

\usepackage[capitalize]{cleveref}
\crefname{section}{Sec.}{Secs.}
\Crefname{section}{Section}{Sections}
\Crefname{table}{Table}{Tables}
\crefname{table}{Tab.}{Tabs.}

\usepackage{newfloat}
\usepackage{listings}
\DeclareCaptionStyle{ruled}{labelfont=normalfont,labelsep=colon,strut=off} 
\floatstyle{ruled}
\newfloat{listing}{tb}{lst}{}
\floatname{listing}{Listing}
\title{Scaling Up Semi-supervised Learning with Unconstrained Unlabelled Data}
\author{
Shuvendu Roy, Ali Etemad
}
\affiliations{ 
Queen's University, Canada\\
\{shuvendu.roy, ali.etemad\}@queensu.ca\\
    
}

\usepackage{bibentry}

\begin{document}

\maketitle
\begin{abstract}
We propose UnMixMatch, a semi-supervised learning framework which can learn effective representations from unconstrained unlabelled data in order to scale up performance. Most existing semi-supervised methods rely on the assumption that labelled and unlabelled samples are drawn from the same distribution, which limits the potential for improvement through the use of free-living unlabeled data. Consequently, the generalizability and scalability of semi-supervised learning are often hindered by this assumption. Our method aims to overcome these constraints and effectively utilize unconstrained unlabelled data in semi-supervised learning. UnMixMatch consists of three main components: a supervised learner with hard augmentations that provides strong regularization, a contrastive consistency regularizer to learn underlying representations from the unlabelled data, and a self-supervised loss to enhance the representations that are learnt from the unlabelled data. We perform extensive experiments on 4 commonly used datasets and demonstrate superior performance over existing semi-supervised methods with a performance boost of 4.79\%. Extensive ablation and sensitivity studies show the effectiveness and impact of each of the proposed components of our method. The code for our work is publicly available\footnote{https://github.com/ShuvenduRoy/UnMixMatch}.
\end{abstract}

\section{Introduction}
Semi-supervised learning (SSL) uses large amounts of \textit{unlabeled} data along with small amounts of \textit{labelled} data to reduce the reliance on fully-labelled datasets. Most existing semi-supervised methods rely on the assumption that the labelled and unlabelled data belong to the same distributions, an assumption that is not necessarily true in real-world scenarios. Moreover, this assumption prohibits us from leveraging free-living unlabelled data with different distributions. In fact, it has been shown in previous studies that incorporating out-of-distribution data with the unlabelled set for SSL impairs performance \cite{oliver2018realistic}.

\begin{figure}
    \centering\includegraphics[width=1\columnwidth]{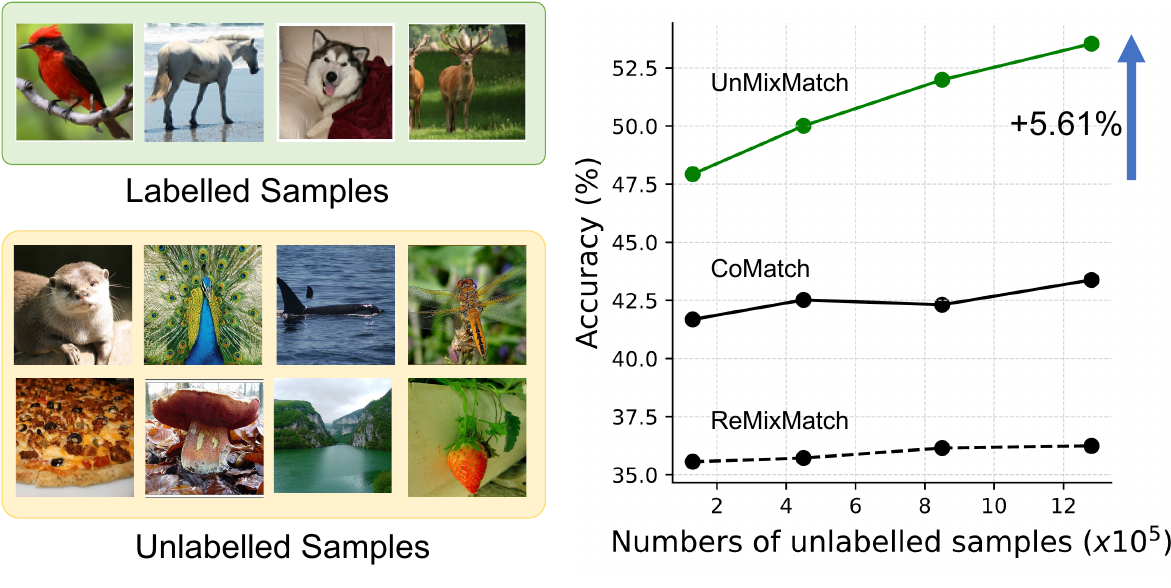}
    \caption{Learning from unconstrained unlabelled data (left). Scaling up unlabelled data provides a large improvement for UnMixMatch (right).}
    \label{fig:bannar}
    
\end{figure}

To adopt a less constrained approach regarding unlabelled data, \textit{open set} SSL has been proposed \cite{openmatch,MTC}, which allows the unlabeled training set to contain samples from classes which are not necessarily present in the labelled set. Nevertheless, this setting still places certain restrictions on the unlabeled data, necessitating the inclusion of samples from \textbf{every} known class and ensuring that its data distribution is similar. These constraints create two important challenges. First, collecting an unlabelled dataset that necessarily includes samples from certain classes can be challenging in real-world settings.
Second, this approach severely restricts the scalability of SSL to large, web-scale, unconstrained, unlabelled data since such data do not hold the aforementioned constraints. Most existing semi-supervised methods are not suitable for learning from unconstrained data as they rely on pseudo-label predictions, which require the  `unlabeled' set to have the same classes as the `labelled' set.

In this paper, we propose a novel SSL approach called UnMixMatch, which can learn effective representations from unconstrained unlabelled data and effectively enable SSL to scale up using web-scale unlabelled data. UnMixMatch comprises three main components, which have some similarities to previous SSL methods but have been specifically modified and tailored toward the `scalability' criteria:
(1) A supervised learner with hard augmentations: We introduce a new hard augmentation module that combines RandAug with MixUp to prevent overfitting on the small labelled set. The convention of using soft augmentations of the existing literature does not perform well when learning from unconstrained unlabelled data. 
(2) A contrastive consistency regularizer: The primary unsupervised learning component of our method involves a contrastive regularizer, which learns the underlying data representations by enforcing the model to produce consistent predictions under strong perturbations. In contrast to existing SSL methods, we do not regularize the class predictions, as the unlabeled set contains unknown classes. Instead, we enforce consistency in the predicted embedding space.
(3) A self-supervised pre-text learning module: To further enhance the learned representations, we include a pre-text task called rotation prediction on the unlabelled data, where the model learns by predicting the degree of rotation applied to each sample.

We conduct extensive research on four common datasets: CIFAR-10, CIFAR-100, SVHN, and STL-10. First, we re-implement and benchmark 13 recent semi-supervised methods with unconstrained unlabelled data, using ImageNet-100 as the unlabelled set. We find that existing methods experience performance degradation in unconstrained settings. In comparison, UnMixMatch outperformed existing methods by an average of 4.79\%. Additionally, UnMixMatch exhibits robust scaling capabilities regarding the size of the unlabelled datasets, as we observe an additional 5.61\% improvement when we increase the unlabelled data size by a factor of 10 (see Fig. \ref{fig:bannar}).
Furthermore, we achieved state-of-the-art results in the open set SSL. Finally, we ablate each component of UnMixMatch and demonstrate the crucial role that each component plays in the performance.

In summary, we make the following contributions:  
\begin{itemize}
    \item 
    We propose a novel semi-supervised method that can learn effective representations from unconstrained unlabelled data for the first time.
    \item 
    We conduct an extensive study to benchmark the performance of existing semi-supervised methods when the unlabelled data are not constrained to match the distribution of the labelled data.
     \item 
     We demonstrate that our method outperforms previous methods by a large margin in unconstrained learning and sets a new state-of-the-art for open set SSL.
    \item 
    We also show that the performance of our method scales up by increasing the amount of unconstrained unlabelled data.
    
    \item To facilitate reproducibility, we release the code at github.com/ShuvenduRoy/UnMixMatch.
\end{itemize}

\section{Related Work}

In this section, we discuss recent developments in SSL in constrained settings, followed by open set SSL.

\subsection{Constrained Semi-supervised Learning}
Prior works on semi-supervised can be broadly divided into two main categories: pseudo-labelling and consistency regularization. Pseudo-labelling techniques \cite{pseudo_labels,noisystudent} mainly rely on the strategy of predicting pseudo-labels for the unlabeled data using the encoder being trained and learns with a combination of the labelled data and unlabeled data plus their pseudo-labels, and iterating on those predictions to gain gradual improvements. Consistency regularization techniques \cite{mean_teacher,pi-model,vat} learn by forcing the embeddings of different augmented unlabeled samples to be similar. This takes place while the model also simultaneously learns via a supervised loss, which is optimized on the labelled samples. In the pseudo-labelling category, Lee et al. \shortcite{pseudo_labels} first introduced the overall approach, and subsequent works improved the technique by adding various interesting elements. For example, in Noisy Student \cite{noisystudent}, a pre-trained teacher was introduced to generate the pseudo-labels, and a student learned from the pseudo-labels along with the labelled data.
In the consistency regularization category, Pi-model \cite{pi-model} was one of the earliest works which used a consistency loss on the predictions of two augmentations of a sample. Later, Mean Teacher \cite{mean_teacher} improved the performance by enforcing consistency between the predictions of an online encoder and an exponential moving average (EMA) encoder. VAT \cite{vat} is another modification of Mean Teacher, which replaced the augmentations with adversarial perturbations. Later, Unsupervised Domain Adaptation (UDA) \cite{uda} showed large improvements by using hard augmentations.

It should be noted that one of the key weaknesses of pseudo-label-based methods is the confidence-bias problem, which arises when the model generates confident wrong pseudo-labels. Yet, their ability to virtually increase the labelled set size by generating pseudo-labels for the unlabeled data has motivated researchers to combine them with consistency regularization methods within the same framework. For instance, MixMatch \cite{mixmatch} predicts pseudo-labels for unlabelled samples while enforcing consistency across augmented images. ReMixMatch \cite{remixmatch} improved upon MixMatch with several implementation tricks, such as augmentation anchoring and distribution alignment. FixMatch is another popular hybrid method known for its simplicity and performance. It predicts the pseudo-label of a sample from a weakly augmented image and treats it as a label for a heavily augmented sample when the confidence of the pseudo-label is above a certain threshold. Subsequently, several works have attempted to improve different aspects of FixMatch. FlexMatch \cite{flexmatch} employs an adaptive curriculum threshold for each class based on that class's performance. CoMatch \cite{comatch} improves upon FixMatch by introducing a graph-based contrastive loss that learns both class representations and low-dimensional embeddings. ConMatch \cite{conmatch} also employs a similar concept, using pseudo-labels as supervision in a contrastive loss. Similarly, SimMatch \cite{simmatch} improves FixMatch by introducing an instance similarity loss in addition to the semantic similarity loss imposed by pseudo-labels. ScMatch \cite{scmatch} utilizes the concept of clustering with SSL by dynamically forming super-classes.

\subsection{Open Set Semi-supervised Learning}   

    Open set SSL is a type of SSL where the unlabelled set includes samples from unknown classes. In prior work \cite{oliver2018realistic}, it was demonstrated that \textit{the presence of unknown classes in the unlabelled dataset has a severe \textbf{negative impact} on the performance of semi-supervised methods}. Similar findings were also reported by Su et al. \shortcite{su2021realistic} that analyzed the performance of more recent semi-supervised methods in open set settings. Nonetheless, some recent methods have attempted to effectively address the detrimental effect of unknown classes while learning from open set unlabelled data. For example, Yoshihashi et al. \shortcite{yoshihashi2019classification} learned to distinguish known classes from unknown ones, effectively avoiding samples from unknown classes in the learning process. A similar approach was taken in Guo et al. \shortcite{guo2020safe}, which proposed a novel scoring function called energy discrepancy to detect and remove instances of unknown classes. OpenMatch \cite{openmatch} used the concept of out-of-distribution to mitigate the negative impact of unknown classes. CCSSL \cite{ccssl} introduced a class-aware contrastive learning approach to improve performance in open set settings. In this study, we tackle a more challenging scenario where the unlabelled set contains instances of unknown classes and does not necessarily include all the known classes. Consequently, our goal is not to detect and remove the images of unknown classes, but rather to learn from them.

\section{Method}

\subsection{Preliminaries and Overview} 
Let, $X_U={(x_i)_{i=1}^N}$ be an unlabelled set, and $X_L={(x_i,y_i)_{i=1}^n}$ be a labelled set where $n \ll N$. In general, semi-supervised methods learn from $X_L$ and $X_U$ in supervised and unsupervised settings, respectively. Formally, SSL is formulated as:
    \begin{equation}
    		\min_{\theta}  
      {\sum_{(x,y)\in X_L}\mathcal{L}_S(x,y,\theta)} 
      +\alpha   
      {\sum_{x\in X_U}\mathcal{L}_U(x,\theta)}
      ,
    		\label{equ: semiLoss}
    	\end{equation}
where, $\theta$ represents the learnable model, $L_S$ is the supervised loss, and $L_U$ is the unsupervised loss. Trivially, it is assumed that $X_L$ and $X_U$ come from the same \textbf{data} and \textbf{class} distributions. Let $Y_l$ and $Y_u$ be the set of classes for labelled and unlabelled data. In the \textit{constrained} setting, $Y_l=Y_u$, and in the \textit{open set} setting, $Y_l$ is a proper subset of $Y_u$, i.e., $Y_l \subset Y_u$. Regarding data distribution, both settings assume that the data comes from the same source and hence have similar distributions. These assumptions are hard to satisfy in a real-world task.

In this study, our objective is to learn from unconstrained data that does not have any particular constraints and can come from different data or class distributions, or both. The unlabelled set may consist of images of unknown classes, where $Y_u \backslash Y_l \neq \varnothing$. Additionally, the unlabelled set may not contain all the known classes, and in the extreme case, $Y_u \cap Y_l = \varnothing$. 
To address this challenge, we propose UnMixMatch, a method that can learn effective visual representations from \textit{\textbf{unconstrained unlabelled}} data. UnMixMatch comprises three modules, which we discuss in detail in the following subsections.

\begin{figure}
    \centering\includegraphics[width=1\columnwidth]{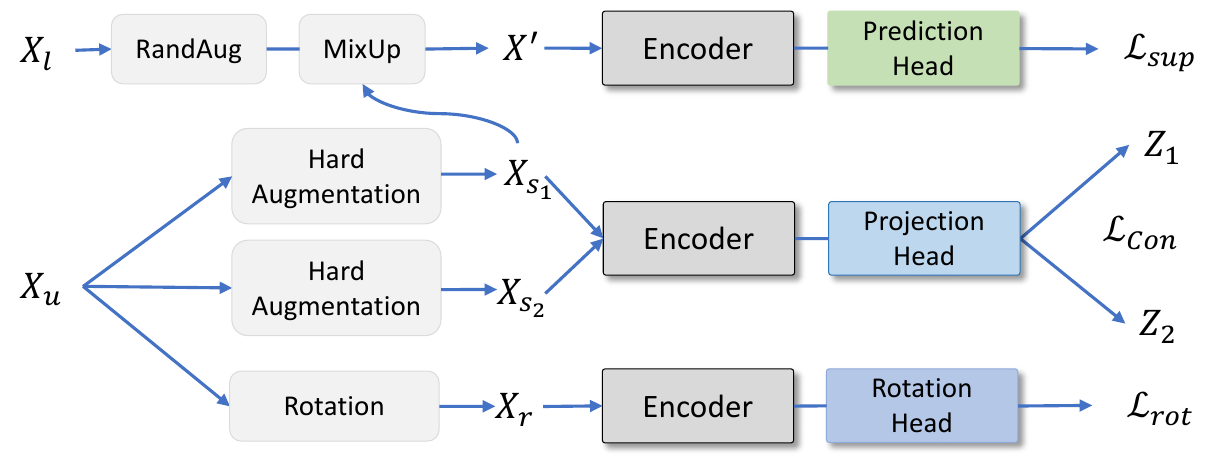}
    \caption{Overview of our proposed method. 
    }
    \label{fig:method}    
\end{figure}

\subsection{Supervised Module}

As mentioned earlier, SSL requires $X_L$ to be learned using a supervised component. The first contribution of our method is, therefore, to create a supervised learner suitable for our purpose of scalable SSL. Here, we hypothesize that given large amounts of unlabeled data in an unconstrained setting and relatively very small amounts of labelled data, the supervised module may overfit the small labelled set. Thus, unlike FixMatch \cite{fixmatch}, MixMatch \cite{mixmatch}, and ReMixMatch \cite{remixmatch}, which use weak augmentations for their supervised modules, we apply hard augmentations on the labelled samples in our supervised module. This acts as a regularizer for supervised learning and is better able to deal with overfitting compared to weaker augmentations. We utilize RandAug \cite{randaugment} as the hard augmentation followed by the MixUp operation \cite{mixmatch,remixmatch}. We denote RandAug plus MixUp as the RandMixUp operation. Finally, a supervised loss is applied to a batch of samples.

\textbf{RandAug.} This is a hard augmentation technique for generating diverse samples by employing a sequence of transformations \cite{randaugment}. More specifically, it applies $R_n\in{1..13}$ transformations randomly chosen from a list of 13 augmentations, including rotation, translation, and colour distortion. The magnitude of each transformation is sampled randomly from a pre-defined range. We denote the augmentation operation as $\hat{x} = RandAug(x)$.

\textbf{MixUp.} 
Let $(x_1, y_1)$ and $(x_2,y_2)$ be two pairs of samples and their class labels. MixUp operation interpolates between the data points to generate mixed samples and labels as:
\begin{equation}
    \Bar{x} = \lambda \cdot {x}_1  + (1 - \lambda) \cdot {x}_2
\end{equation}
\begin{equation}
    \Bar{y} = \lambda \cdot y_1  + (1 - \lambda) \cdot y_2 ,
\end{equation}
where $\lambda$ is the mixing coefficient. Following MixMatch \cite{mixmatch}, we sample $\lambda$ from a beta distribution $(\lambda \sim Beta(\alpha, \alpha))$ with hyper-parameter $\alpha$.

\textbf{Supervised Loss.}
For a batch of unlabeled samples $X_u=((\hat{x}_i); i\in(1,..,b))$, with batch size $b$, we first generate the pseudo-label for each sample $X_i$ as
$p_i = P_\theta(x_i)$,
where $P_\theta$ is the encoder with a classification head.

Next, for a batch of labelled samples $X_l=((\hat{x}_i, y_i); i\in(1,...,b))$ and the unlabelled samples with pseudo-labels $X_p=((\hat{x}_i, p_i); i\in(1,..,b))$, we augment all the samples using $\Bar{X} = RandAugMix(X_l, X_p)$. 
Accordingly, we define the supervised loss of our method as: 
\begin{align}
    \mathcal{L}_{sup} = \frac{1}{b} \sum_{\Bar{x},\Bar{y}\in \Bar{X}} \mathcal{H}(\Bar{y}, P_\theta(y | \Bar{x})),
\end{align}
where $\mathcal{H}$ is the cross-entropy loss.

\subsection{Consistency Regularization Module}
To deal with the unconstrained nature of existing unlabeled data and learn effective representations, we apply a consistency regularizer as our second contribution.
A consistency regularizer learns from the unlabelled data by enforcing consistency on its predictions under different augmentations. 
Prior works that have used consistency regularization for SSL \cite{pi-model,mixmatch,remixmatch} enforce consistency on the \textbf{class predictions} under different perturbations. However, regularization over class predictions is not useful for learning in \textit{unconstrained} settings where unlabelled data do not necessarily come from the same classes as the labelled data. To address this, we enforce consistency in the low-dimensional embedding space using a contrastive loss. Using contrastive loss on the embedding space enables the model to learn class-agnostic representations from unconstrained unlabelled data. In UnMixMatch, we adopt the Noise Contrastive Estimation loss, a.k.a InfoNCE \cite{simclr}. 

Contrastive learning learns from positive (perturbations of the same sample) and negative samples (all other samples) by bringing embeddings of the positive samples together and pushing them away for the negatives. For each sample, $x_i\in X_U$, two augmentations are applied to generate two augmented images $\hat{x_i} = RandAug(x_i)$. These are first passed through the encoder and a projection head (shallow linear layers with non-linearity and batch normalization) to obtain embeddings $z_i = P_{\theta_p}(z | \hat{x_i})$. The contrastive loss is accordingly defined as:
\begin{equation}
\label{eq:loss_self}
    \mathcal{L}_{\text{\textit{con}}} = - \frac{1}{2b} \sum_{i=1}^{2b} log\frac{exp(z_i, z_{\kappa(i)}/\tau)}{\sum_{k=1}^{2b} \mathbf{1}_{[k \neq i]} exp(z_i, z_k/\tau)} ,
\end{equation}
where, $\kappa(i)$ is the index of the second augmented sample, $\mathbf{1}_{[k \neq i]}$ is an indicator function which returns 1 when $k$ is not equal to $i$, and 0 otherwise. $\tau$ is a temperature parameter.

\subsection{Self-supervised Module}
Finally, we intend to enhance the quality of the representations extracted from the unconstrained unlabelled data using the consistency regularizer. It has been recently shown that self-supervised techniques can be employed to learn underlying domain-invariant representations for unlabelled data \cite{rotation,colorization}. Moreover, this idea has already been demonstrated to be useful in conjunction with SSL \cite{rotation,s4l,remixmatch}. As a result, we integrate a straightforward yet effective self-supervised pre-text task called rotation prediction, which learns by predicting the degree of rotations applied to unlabelled images. In practice, a rotation module randomly samples one of the following rotations and applies it to an unlabelled image: {$0^{\circ}, 90^{\circ},180^{\circ},270^{\circ}$}. As a result, the rotation prediction task can be viewed as a four-way classification problem, represented as: 
\begin{align} 
    \mathcal{L}_{rot} = \frac{1}{b} \sum_{u\in U} \mathcal{H}(r, P_{\theta_r}(r | Rotate(u))
\end{align}
Here, $P_{\theta_r}$ is the encoder with a rotation head that predicts the rotation, and $\mathcal{H}$ is the cross-entropy loss.

{\renewcommand{\arraystretch}{1.0}   
\begin{table*}[t!]
\centering

\setlength{\tabcolsep}{3.5pt}
\resizebox{1\textwidth}{!}
{
    \begin{tabular}{l|ccc|ccc|ccc|c | c}
    	& \multicolumn{3}{c|}{\textbf{CIFAR-10}}& \multicolumn{3}{c|}{\textbf{CIFAR-100}}& \multicolumn{3}{c|}{\textbf{SVHN}}& \multicolumn{1}{c|}{\textbf{STL-10} } & 
         \\
        \hline
        \textbf{Methods} & 40 labels & 250 labels & 4000 labels& 400 labels & 2500 labels & 10000 labels& 40 labels & 250 labels & 1000 labels & 1000 labels & \textbf{Avg.}\\ \shline

Supervised & 24.24\scriptsize{±1.1} & 	 43.33\scriptsize{±2.2} & 	 83.76\scriptsize{±0.4} & 	 10.39\scriptsize{±0.3} & 	 39.57\scriptsize{±0.4} & 	 63.4\scriptsize{±0.1} & 	 24.67\scriptsize{±2.1} & 	 75.42\scriptsize{±2.4} & 	 87.74\scriptsize{±0.6} & 	 60.96\scriptsize{±1.3} &51.35 \\
Pi-Model \cite{pi-model} & 24.16\scriptsize{±1.5} & 	 48.14\scriptsize{±1.1} & 	 84.19\scriptsize{±0.3} & 	 12.81\scriptsize{±0.9} & 	 40.01\scriptsize{±0.3} & 	 63.55\scriptsize{±0.1} & 	 28.02\scriptsize{±1.1} & 	 76.56\scriptsize{±2.2} & 	 87.83\scriptsize{±0.6} & 	 69.73\scriptsize{±0.6} &53.5 \\
Mean Teacher \cite{mean_teacher} & 26.22\scriptsize{±1.0} & 	 49.35\scriptsize{±1.7} & 	 83.7\scriptsize{±0.4} & 	 13.72\scriptsize{±0.9} & 	 41.57\scriptsize{±0.3} & 	 63.83\scriptsize{±0.1} & 	 28.57\scriptsize{±1.3} & 	 76.78\scriptsize{±2.3} & 	 87.68\scriptsize{±0.6} & 	 70.12\scriptsize{±0.1} &54.15 \\
VAT \cite{vat} & 24.7\scriptsize{±1.3} & 	 46.18\scriptsize{±1.3} & 	 84.73\scriptsize{±0.2} & 	 11.5\scriptsize{±0.8} & 	 41.73\scriptsize{±0.2} & 	 63.76\scriptsize{±0.1} & 	 41.95\scriptsize{±2.5} & 	 76.18\scriptsize{±1.9} & 	 88.07\scriptsize{±0.5} & 	 63.12\scriptsize{±0.6} &54.19 \\
Pseudo-label \cite{pseudo_labels} & 24.88\scriptsize{±1.6} & 	 50.29\scriptsize{±1.3} & 	 84.11\scriptsize{±0.1} & 	 12.12\scriptsize{±0.1} & 	 39.72\scriptsize{±0.8} & 	 63.57\scriptsize{±0.0} & 	 36.04\scriptsize{±2.7} & 	 78.04\scriptsize{±1.2} & 	 88.91\scriptsize{±0.3} & 	 65.6\scriptsize{±0.9} &54.33 \\
UDA \cite{uda} & 28.12\scriptsize{±2.5} & 	 65.59\scriptsize{±1.7} & 	 88.31\scriptsize{±0.1} & 	 21.11\scriptsize{±0.8} & 	 51.82\scriptsize{±0.6} & 	 69.42\scriptsize{±0.6} & 	 48.8\scriptsize{±4.1} & 	 77.73\scriptsize{±1.9} & 	 88.83\scriptsize{±0.3} & 	 82.54\scriptsize{±0.2} &62.23 \\
MixMatch \cite{mixmatch} & 32.58\scriptsize{±0.8} & 	 58.24\scriptsize{±0.3} & 	 84.59\scriptsize{±0.3} & 	 20.26\scriptsize{±0.6} & 	 45.94\scriptsize{±0.4} & 	 65.89\scriptsize{±0.2} & 	 57.46\scriptsize{±1.7} & 	 77.65\scriptsize{±1.2} & 	 89.95\scriptsize{±0.2} & 	 71.85\scriptsize{±0.6} &60.44 \\
ReMixMatch \cite{remixmatch} & 35.56\scriptsize{±0.8} & 	 64.71\scriptsize{±0.6} & 	 87.64\scriptsize{±0.3} & 	 18.9\scriptsize{±0.8} & 	 49.11\scriptsize{±0.9} & 	 69.38\scriptsize{±0.1} & 	 56.94\scriptsize{±4.2} & 	 79.57\scriptsize{±0.9} & 	 \underline{90.57\scriptsize{±0.3} } & 	 79.13\scriptsize{±0.8} &63.15 \\
FixMatch \cite{fixmatch} & 27.91\scriptsize{±3.9} & 	 64.98\scriptsize{±1.0} & 	 88.18\scriptsize{±0.1} & 	 21.2\scriptsize{±0.6} & 	 51.4\scriptsize{±1.3} & 	 67.8\scriptsize{±0.3} & 	 43.71\scriptsize{±5.5} & 	 74.52\scriptsize{±1.0} & 	 88.06\scriptsize{±0.6} & 	 81.85\scriptsize{±0.4} &60.96 \\
FlexMatch \cite{flexmatch} & 32.8\scriptsize{±0.8} & 	 63.22\scriptsize{±1.0} & 	 87.82\scriptsize{±0.1} & 	 20.87\scriptsize{±0.7} & 	 51.28\scriptsize{±0.8} & 	 69.52\scriptsize{±0.5} & 	 \underline{60.5\scriptsize{±2.5} } & 	 79.72\scriptsize{±0.6} & 	 88.85\scriptsize{±0.4} & 	 \underline{82.67\scriptsize{±0.4} } &63.72 \\
CoMatch \cite{comatch} & \underline{41.68\scriptsize{±0.7} } & 	 62.31\scriptsize{±0.7} & 	 84.52\scriptsize{±0.3} & 	 22.6\scriptsize{±0.6} & 	 44.0\scriptsize{±1.0} & 	 58.55\scriptsize{±0.3} & 	 45.87\scriptsize{±2.8} & 	 73.19\scriptsize{±0.3} & 	 86.45\scriptsize{±0.2} & 	 82.0\scriptsize{±0.0} &60.12 \\
CCSSL \cite{ccssl} & 30.89\scriptsize{±5.9} & 	 \underline{67.2\scriptsize{±1.5} } & 	 \underline{88.77\scriptsize{±0.1} } & 	 \underline{24.53\scriptsize{±1.5} } & 	 \textbf{56.3\scriptsize{±0.2} } & 	 \underline{71.13\scriptsize{±0.3} } & 	 50.02\scriptsize{±6.6} & 	 \underline{80.39\scriptsize{±0.6} } & 	 88.6\scriptsize{±0.3} & 	 82.0\scriptsize{±0.0} &\underline{63.98} \\ 
SimMatch \cite{simmatch} & 23.77\scriptsize{±1.8} & 	 57.72\scriptsize{±1.3} & 	 84.12\scriptsize{±0.7} & 	 18.65\scriptsize{±1.2} & 	 47.33\scriptsize{±1.0} & 	 66.54\scriptsize{±0.8} & 	 51.23\scriptsize{±1.6} & 	 74.48\scriptsize{±1.1} & 	 88.57\scriptsize{±1.0} & 	 77.23\scriptsize{±1.2} &58.96 \\
ScMatch \cite{scmatch} & 27.81\scriptsize{±1.1} & 	 56.78\scriptsize{±0.6} & 	 83.09\scriptsize{±0.2} & 	 18.14\scriptsize{±1.1} & 	 46.21\scriptsize{±0.7} & 	 64.24\scriptsize{±0.2} & 	 56.59\scriptsize{±0.9} & 	 75.08\scriptsize{±0.8} & 	 89.23\scriptsize{±0.3} & 	 79.44\scriptsize{±0.7} &59.66 \\ 
UnMixMatch & \textbf{47.93\scriptsize{±1.1} } & 	 \textbf{68.72\scriptsize{±0.6} } & 	 \textbf{89.58\scriptsize{±0.2} } & 	 \textbf{26.13\scriptsize{±1.1} } & 	 \underline{54.18\scriptsize{±0.7} } & 	 \textbf{71.73\scriptsize{±0.2} } & 	 \textbf{72.9\scriptsize{±0.9} } & 	 \textbf{80.78\scriptsize{±0.8} } & 	 \textbf{91.03\scriptsize{±0.3} } & 	 \textbf{84.73\scriptsize{±0.7} } &\textbf{68.77} \\
    \end{tabular}
    
}
\caption{
Comparison of our method against other SSL methods with \textit{unconstrained unlabelled data} on 4 different datasets.} 
\label{tab:ssl_final}

\end{table*}
}

{\renewcommand{\arraystretch}{1.0}   
\begin{table}[t!]
\centering

\setlength{\tabcolsep}{3pt}
\resizebox{0.75\columnwidth}{!}
{
    \begin{tabular}{l c c c c}
    \textbf{Data} &\textbf{IN-100} & \textbf{Subset 1}& \textbf{Subset 2}& \textbf{IN-1K}\\
    \hline
    No. of samples & 130K & 450K & 850K & 1.28M \\
    \shline
    ReMixMatch & 35.56 & 35.72 & 36.15 &  36.24\\ 
    CoMatch & 41.68 &  42.52 &  42.31 &  43.38 \\ 
UnMixMatch & \textbf{47.93} & \textbf{50.01} & \textbf{51.99} & \textbf{53.54} \\ 
\end{tabular}
    
}
\caption{The impact of unlabeled set size. 
Here, Subsets 1 \& 2 are two random subsets of ImageNet-1K (IN-1K). } 
\label{tab:more_unlabelled_data}
\end{table}
}

\subsection{Total Loss}
Finally, we incorporate the loss functions for the three modules above to create the total loss: 
\begin{equation}
    \mathcal{L}_{UnMixMatch} = \mathcal{L}_{sup} + \beta \mathcal{L}_{con} + \gamma\mathcal{L}_{rot}.
\end{equation}
Here, $\beta$ and $\gamma$ are hyper-parameters that balance the significance of the contrastive and rotation losses. An overview of our proposed method is presented in Fig. \ref{fig:method}.

\section{Experiments and Results}

\subsection{Datasets and Implementation Details}\label{sec:res-implementation}
For our main experiments, we follow the standard semi-supervised evaluation protocol from prior works \cite{fixmatch}, and present the results for four datasets: CIFAR-10 \cite{krizhevsky2009learning}, CIFAR-100 \cite{krizhevsky2009learning}, SVHN \cite{netzer2011reading}, and STL-10 \cite{coates2011analysis}. We present the results for different numbers of labelled samples, averaged over three runs. We use ImageNet-1K \cite{imagenet}, and ImageNet-100 (a subset of ImageNet-1K) as the unconstrained unlabeled datasets since it has different class and data distribution in comparison to the four aforementioned datasets.

Our implementation and hyper-parameters closely follow FlexMatch \cite{flexmatch}. For the encoder, following existing literature such as \cite{fixmatch,flexmatch}, we use Wide ResNet-28-2 \cite{wideresnet} for CIFAR-10 and SVHN, WRN-28-8 for CIFAR-100, and WRN-37-2 for STL-10. We train the method for $2^{20}$ iterations with a batch size of 64, a learning rate of 0.03, and an SGD optimizer with a momentum of 0.9 and weight-decay of 0.0005. The code is implemented with Pytorch and built using TorchSSL \cite{flexmatch}.

{\renewcommand{\arraystretch}{1.0}   
\begin{table}[t!]
\centering

\setlength{\tabcolsep}{4.5pt}
\resizebox{0.48\textwidth}{!}
{
    \begin{tabular}{l ccc }
    & \multicolumn{3}{c}{\textbf{Labelled samples/class}} \\ 
    \hline
     \textbf{Methods}
     & \textbf{50} & \textbf{100} & \textbf{400} \\
     \shline 
     Supervised & 64.3\scriptsize{±1.1} & 69.5\scriptsize{±0.7}& 80.0\scriptsize{±0.3 } \\
     FixMatch \cite{fixmatch} & 56.8\scriptsize{±1.2} & 70.2\scriptsize{±0.6}& 83.7\scriptsize{±0.5 } \\
     MTC \cite{MTC} & 79.7\scriptsize{±0.9} & 86.3\scriptsize{±0.9}& 91.0\scriptsize{±0.5 } \\
     OpenMatch \cite{openmatch} & 89.6\scriptsize{±0.9} & 92.9\scriptsize{±0.5}& 94.1\scriptsize{±0.5 } \\ 
    UnMixMatch & \textbf{95.7\scriptsize{±0.8}} & \textbf{96.8\scriptsize{±0.5}} & \textbf{97.2\scriptsize{±0.4 }} \\
    \end{tabular}   
}
\caption{Performance comparison on open set SSL for CIFAR-10 with 6/4 known-unknown class split.} 
\label{tab:open_set_ssl}
\end{table}
}

\subsection{Results}\label{sec:result}
Here, we present the main results of our method, including the performance on the four datasets (with ImageNet-100 for unlabeled data) in unconstrained settings, analysis of UnMixMatch's performance with increasing the number of unlabelled data, its effectiveness in open set settings, and its performance in a barely supervised setting.

\subsubsection{Unconstrained Settings.}\label{sec:res-main}
Table \ref{tab:ssl_final} presents the main results of our work on the four datasets.
Here, we first re-implement 13 semi-supervised methods and report the results with \textit{unconstrained} unlabelled data.
We report the average accuracies and standard deviations across three individual runs for each setting. We also report the average accuracy across all settings for overall comparison. 
It should be noted that the performance of prior methods is considerably lower than what has been reported in the original papers, where the unlabeled and labelled samples came from the same datasets (unlabeled data were not unconstrained).
Next, we observe that UnMixMatch demonstrates superior performance compared to other methods, with an average improvement of 4.79\%. We obtain considerable improvement across all datasets and splits, except using CIFAR-100 with 2500 labels, where CCSSL achieves a better result. 

When considering the number of labelled samples, we notice that the differences between UnMixMatch and other methods are more pronounced when the labelled set size is small. For example, with only 40 labelled samples from CIFAR-10, UnMixMatch achieves a 17.04\% performance gain over CCSSL (which has the second highest overall average performance after ours), and 6.25\% higher than the next best result for this specific setting, which was obtained by CoMatch. 
A similar pattern is observed for SVHN, where UnMixMatch outperforms CCSSL and FlexMatch by 22.88\% and 12.4\%, respectively.

\subsubsection{Scaling Up the Unlabelled Set.}\label{sec:res-scalingup}
Our main motivation for using unconstrained unlabelled data is to take advantage of the abundance of free-living unlabeled data. 
In this experiment, we evaluate the performance of UnMixMatch as the size of the unlabelled set is increased. The results of this study are summarized in Fig. \ref{fig:bannar} and presented in detail in Table \ref{tab:more_unlabelled_data}. Specifically, we increase the size of the unlabelled set from 130K images of ImageNet-100 (a subset of ImageNet-1K) to 1.28M images of ImageNet-1K, with two more subsets of 450K and 850K images from ImageNet-1K. We perform this experiment on CIFAR-10 with 40 labelled samples with the two best methods (CoMatch and ReMixMatch) on this setting and observe an increasing trend in the accuracy of UnMixMatch as the number of images in the unlabelled set increases. 
With ImageNet-1K used as the unlabelled set, which is approximately 10 times larger than ImageNet-100, the accuracy of UnMixMatch improves from 47.93\% to 53.54\%, a significant improvement of 5.61\% by simply increasing the size of the unlabelled set. CoMatch and ReMixMatch, on the other hand, show very small improvements with the increase in the unlabelled data, but the performance difference with our method further increases. 

{\renewcommand{\arraystretch}{1.0}   
\begin{table}[t!]
\centering

\setlength{\tabcolsep}{9.5pt}
{
    \small
    \begin{tabular}{l c }
    \textbf{Method} & \textbf{Accuracy} (\%)  \\
    \shline
    FlexMatch \cite{flexmatch} & 21.96\scriptsize{±1.4}   \\
    CCSSL \cite{ccssl} & 15.63\scriptsize{±1.7}  \\
    \textbf{UnMixMatch} &\textbf{ 27.54\scriptsize{±2.5} }\\
\end{tabular}
    
}
\caption{Performance comparison on barely supervised learning. Only 1 sample per class is used for training.} 
\label{tab:barely_supervised_ssl}
\end{table}
}

\subsubsection{Results on Open Set Setting.}\label{sec:res-openset}
Next, we investigate the performance of UnMixMatch on open set SSL. Open set SSL is a relatively less challenging setting than \textit{unconstrained} settings, where the unlabelled set may contain images of unknown classes but must contain images of all known classes \cite{openmatch,MTC}. 
For learning in open set settings, we employ a variant of UnMixMatch which takes advantage of the fact that the unlabelled set contains samples of all known classes from the labelled set and learns from the predicted pseudo-labels on the unlabelled set. In this variant, we replace the contrastive loss in our method with the class-aware contrastive loss of CCSSL \cite{ccssl}. This method first predicts the pseudo-labels for the unlabelled samples and uses them with contrastive loss to learn clusters of known classes in the embedding space. For this experiment, we follow the experimental setup of OpenMatch \cite{openmatch}, which reports the results for CIFAR-10 with a 6/4 split. This split means that the labelled set contains images of 6 classes from CIFAR-10, while the unlabelled set includes images of 6 known and 4 unknown classes. Like OpenMatch, we take 6 animal classes as the known classes and 4 object classes as the unknown classes. We perform this experiment using three different splits with 50, 100, and 400 labelled samples per class in the known set. The results of this experiment are presented in Table \ref{tab:open_set_ssl}, where it can be observed that our method outperforms the existing methods and sets a new state-of-the-art for \textit{open set SSL}. Once again, UnMixMatch better demonstrates its effectiveness when the amount of labelled data is limited. With 50 labelled samples per class, our approach provides a 6.1\% improvement over the second-best method, OpenMatch. For 100 and 400 labelled samples per class, UnMixMatch shows 3.9\% and 3.1\% improvements, respectively.

\subsubsection{Barely Supervised Setting.}\label{sec:res-barely}
In this section, we aim to test the performance of SSL in the extreme scenario where only one labelled sample per class is available. This \textit{barely supervised} setting is considered to be very challenging, even with conventional SSL techniques that use \textit{constrained} unlabelled data. Given the extremely low number of labelled samples, the results in such settings generally exhibit high variance, and therefore, we increase the number of folds to 5 to account for this variability. This is due to the fact
that the quality of labelled data greatly influences the performance in such low data settings \cite{fixmatch,roy2022impact}. The results of this experiment are presented in Table \ref{tab:barely_supervised_ssl}. It shows the performance of CIFAR-10 for the three best methods identified previously in Table \ref{tab:ssl_final}.

In this experiment, UnMixMatch achieves an accuracy of 27.54\% and outperforms other methods by 5.58\%. 
As before, CCSSL struggles to learn in low data settings, achieving a near chance-level accuracy of only 15.63\%. However, FlexMatch shows relatively better performance with an accuracy of 21.96\%.  In general, we find that it is quite difficult to learn effective representations with just one labelled sample per class while using unconstrained unlabelled data. However, as previously shown in Table \ref{tab:ssl_final}, UnMixMatch quickly gains significant improvements with a small increase in the number of labelled data and achieves an accuracy of 47.93\% with four labelled samples per class (40 labels in total).

\begin{table*}[t]
\centering
\subfloat[
Ablation of main components.
\label{tab:ablation_main}
]{
\centering
\begin{minipage}{0.29\linewidth}{\begin{center}
\tablestyle{2pt}{1.1}
\begin{tabular}{y{80}x{40}}
Ablation & Accuracy \\
\shline
        UnMixMatch & \textbf{47.93} \\ 
        w/o RandMixUp & 38.72 \\
        w/o Contrast Loss & 41.25 \\ 
        w/o Rotation Loss & 41.83 \\
        \\
        
\end{tabular}
\end{center}}\end{minipage}
}
\hspace{2em}
\subfloat[
Alternate hard augmentations.
\label{tab:ablation:augmentation}
]{
\begin{minipage}{0.29\linewidth}{\begin{center}
\tablestyle{1pt}{1.1}
\begin{tabular}{y{70}x{36}}
Augmentation & Accuracy \\
\shline
    RandMixUp &  47.93    \\ 
    RandAug & 46.01\\
    MixUp & 45.23 \\
    CutMix & 43.12 \\
    AugMix& 44.75\\ 
    
\end{tabular}
\end{center}}\end{minipage}
}
\centering
\hspace{2em}
\subfloat[
Variants of contrastive regularizers.
\label{tab:ablation:contrast}
]{
\begin{minipage}{0.29\linewidth}{\begin{center}
\tablestyle{2pt}{1.1}
\begin{tabular}{y{90}x{45}}
Contrastive loss & Accuracy \\
\shline
Ours &  47.93 \\ 
ConMatch  & 45.82\\
Contrastive Reg. & 45.77\\
Graph Contrast & 47.12\\
Class-aware Contrast & 47.88\\ 
\end{tabular}
\end{center}}\end{minipage}
}
\\ 
\vspace{0.5em}
\subfloat[
Alternate regularization strategies.
\label{tab:ablation:contrast_aug}
]{
\centering
\begin{minipage}{0.29\linewidth}{\begin{center}
\tablestyle{4pt}{1.1}
\begin{tabular}{y{85}x{40}}
Regularization strategy & Accuracy \\
\shline
Cont. two strong (ours) & 47.93\\
Cont. on prediction & 42.45 \\
Cont. weak \& strong & 44.15 \\
\\
\end{tabular}
\end{center}}\end{minipage}
}
\hspace{2em}
\subfloat[
Alternate contrastive regularizers.
\label{tab:ablation:self-sup}
]{
\centering
\begin{minipage}{0.29\linewidth}{\begin{center}
\tablestyle{4pt}{1.1}
\begin{tabular}{y{60}x{45}}
Self-sup. loss & Accuracy \\
\shline
Ours &  47.93 \\ 
BYOL & 46.89\\
SimSiam & 46.50\\
VICReg & 47.91\\
\end{tabular}
\end{center}}\end{minipage}
}
\hspace{2em}
\subfloat[
Alternate pre-text tasks.
\label{tab:ablation:pre-text}
]{
\begin{minipage}{0.29\linewidth}{\begin{center}
\tablestyle{4pt}{1.1}
\begin{tabular}{y{60}x{36}}
Method & Accuracy \\
\shline
    Rot. Pred. & 47.93   \\
    Colorization  & 44.25\\ 
    Jigsaw Solving  & 43.98\\ 
    Image Inpaint & 45.56\\
\end{tabular}
\end{center}}\end{minipage}
}
\caption{Ablation studies on our method. All studies are on CIFAR-10 with 40 labelled samples.
}
\label{tab:ablations}
\end{table*}

\subsection{Ablation Study}\label{sec:ablation} 
\subsubsection{Main Components.}\label{sec:ablation-main}
Table \ref{tab:ablation_main} presents the main ablation results by removing each of the three main components of the proposed method: RandMixUp augmentation, consistency regularization, and rotation prediction. Note that we can not remove two components simultaneously, since semi-supervised learners require a minimum of one supervised and one unsupervised loss. \textcolor{black}{Therefore, we can not perform ablation studies with only one component at a time.} These experiments are done on CIFAR-10 with 40 samples. The table demonstrates that all three components have a significant impact on the final performance of the model, with the removal of any one component resulting in a large drop in accuracy. In the first ablation experiment, we remove the RandMixUp augmentation module, effectively learning from the labelled samples with weak augmentations only (random resized crop and horizontal flip). This experiment results in the highest drop in accuracy across the ablation settings, with a 9.21\% decrease. The second largest drop in accuracy is observed when removing consistency regularization, resulting in a 6.68\% decrease. Similarly, removing the rotation prediction component results in a 6.1\% drop in performance.

\subsubsection{RandMixUp.}\label{sec:ablation-mixup}
As demonstrated in the main ablation study, the RandMixUp augmentation is a crucial component of our approach. In Table \ref{tab:ablation:augmentation}, we present the accuracy of CIFAR-10 with 40 samples for different alternatives to RandMixUp. Recall that RandMixUp combines RandAug \cite{randaugment} and MixUp \cite{mixmatch}. We first test the accuracy for RandAug and MixUp individually and observe a decrease in performance for both settings alone, with a 1.92\% and 2.7\% decrease, respectively. Here, using RandAug exhibits a lower drop than MixUp, suggesting the higher importance of RandAug in the RandMixUp. This analysis also shows that, unlike MixMatch and ReMixMatch, learning using MixUp does not have the most significant impact on the supervised component. Instead, the role of hard augmentation as a regularizer is the key to the impact of RandMixUp in our method.

We also examine the performance of other well-known hard augmentation techniques, namely CutMix and AugMix. Like MixUp, CutMix combines two samples, but in this case, a part of the second sample is cut and inserted into the first sample to create a new one. Despite the conceptual similarity, CutMix yields 2.11\% worse accuracy than MixUp, which is 4.81\% lower than the accuracy achieved by the proposed RandAugMix. On the other hand, AugMix shares a similar concept as RandAug. While RandAug applies multiple augmentations in sequence, AugMix applies multiple augmentations separately and generates the final sample by interpolating between them. AugMix achieves an accuracy of 44.75\%, which is 1.26\% lower than that of RandAug and 3.18\% lower than that of RandAugMix. The study's overall findings demonstrate the critical importance of RandAugMix in our method, with other conceptually similar augmentation techniques failing to achieve the same level of performance.

\subsubsection{Consistency Regularizer.}\label{sec:ablation-contrast}
As shown in the main ablation study, the contrastive regularizer is the second most crucial component of our method. Recall that the proposed contrastive loss for UnMixMatch is a noise contrastive estimation loss that has gained popularity in the self-supervised learning literature \cite{infonce, simclr}. However, other variants of contrastive loss have also been introduced in the literature \cite{conmatch, CR, comatch, ccssl} and have shown improvements in different problem settings, including SSL. In this study, we investigate four variants of contrastive loss in the proposed framework. More specifically, we explore the contrastive variants of ConMatch \cite{conmatch} and Contrastive Regularization \cite{CR}, as well as graph contrastive learning \cite{comatch}, and class-aware contrastive learning \cite{ccssl}. ConMatch \cite{conmatch} is based on a variant of contrastive loss that involves two hard augmentations and one weak augmentation. In Contrastive Regularization \cite{CR}, high-confidence pseudo-labels were used for supervision while learning with a contrastive loss. Two other methods proposed slightly different versions of utilizing the pseudo-labels in contrastive learning settings: CoMatch \cite{comatch} and CCSSL \cite{ccssl}, where CoMatch used a graph contrastive learning and CCSSL utilized pseudo-labels with a contrastive loss for out-of-distribution learning.

We show the results for variants of contrastive loss in Table \ref{tab:ablation:contrast}.
Using the contrastive variant of both ConMatch and Constartive Regularizer result in a large drop in accuracy with 2.16\% and 2.11\%, respectively. Using the graph contract concept of CoMatch with UnMixMatch results in a 47.12\% accuracy, dropping by 0.81\%. Finally, the class-aware contrastive loss of CCSSL gets the closest accuracy of the proposed UnMixMatch with 47.88\% accuracy. In Table \ref{tab:contrast}, we present expanded results across all datasets for the graph contrast from CoMatch \cite{comatch} and the class-aware contrast from CCSSL \cite{ccssl}. Results in this table show that on average the proposed contrastive regularizer achieves 2.99\% and 4.45\% improvements over the graph contrast and class-aware contrast alternatives. The class-aware contrast obtains better results than the proposed method in one setting (CIFAR-100 with 400 labelled samples) only.

{\renewcommand{\arraystretch}{1.0}   
\begin{table*}[t!]
\centering

\setlength{\tabcolsep}{3.5pt}
\resizebox{1\textwidth}{!}
{
    \begin{tabular}{l|ccc|ccc|ccc|c | c}
    	& \multicolumn{3}{c|}{\textbf{CIFAR-10}}& \multicolumn{3}{c|}{\textbf{CIFAR-100}}& \multicolumn{3}{c|}{\textbf{SVHN}}& \multicolumn{1}{c|}{\textbf{STL-10} } & 
         \\
        \hline
        \textbf{Methods} & 40 labels & 250 labels & 4000 labels& 400 labels & 2500 labels & 100000 labels& 40 labels & 250 labels & 1000 labels & 1000 labels & \textbf{Avg.}\\ \hline

Ours & \textbf{47.93\scriptsize{±1.1} } & 	 \textbf{68.72\scriptsize{±0.6} } & 	 \textbf{89.58\scriptsize{±0.2} } & 	 {26.13\scriptsize{±1.1} } & 	 \textbf{54.18\scriptsize{±0.7} } & 	 \textbf{71.73\scriptsize{±0.2} } & 	 \textbf{72.9\scriptsize{±0.9} } & 	 \textbf{80.78\scriptsize{±0.8} } & 	 \textbf{91.03\scriptsize{±0.3} } & 	 \textbf{84.73\scriptsize{±0.7} } & \textbf{68.77} \\

Graph Contrast & {47.12\scriptsize{±0.5} } & 67.67\scriptsize{±0.7} & 	 87.3\scriptsize{±0.3} & 	 25.38\scriptsize{±0.4} & 	 49.14\scriptsize{±0.6} & 	 68.08\scriptsize{±0.3} & 	 62.54\scriptsize{±3.6} & 	 {79.78\scriptsize{±0.9} } & 	 88.49\scriptsize{±0.9} & 	 	 82.25\scriptsize{±0.5} & 65.78  \\

Class-aware Con.  & 47.88\scriptsize{±2.4} & 	 68.45\scriptsize{±0.4} & 	 88.39\scriptsize{±0.2} & 	 \textbf{26.59\scriptsize{±0.3} } & 	 50.25\scriptsize{±0.7} & 	 {69.29\scriptsize{±0.2} } & 43.99\scriptsize{±12.2} & 	 78.05\scriptsize{±0.2} & 	 87.76\scriptsize{±0.1} & 	  	 82.59\scriptsize{±0.4} & 64.32 \\
    \end{tabular}
    
}
\caption{
Comparison of our method against different variants of contrastive regularizers on 4 different datasets.} 
\label{tab:contrast}
\end{table*}
}  

{\renewcommand{\arraystretch}{1.0}   
\begin{table*}[t!]
\centering

\setlength{\tabcolsep}{3.5pt}
\resizebox{1\textwidth}{!}
{
    \begin{tabular}{l|ccc|ccc|ccc|c | c}
    	& \multicolumn{3}{c|}{\textbf{CIFAR-10}}& \multicolumn{3}{c|}{\textbf{CIFAR-100}}& \multicolumn{3}{c|}{\textbf{SVHN}}& \multicolumn{1}{c|}{\textbf{STL-10} } & 
         \\
        \hline
        \textbf{Methods} & 40 labels & 250 labels & 4000 labels& 400 labels & 2500 labels & 100000 labels& 40 labels & 250 labels & 1000 labels & 1000 labels & \textbf{Avg.}\\ \hline

Ours & \textbf{47.93\scriptsize{±1.1} } & 	 {68.72\scriptsize{±0.6} } & 	 \textbf{89.58\scriptsize{±0.2} } & 	 \textbf{26.13\scriptsize{±1.1} } & 	 \textbf{54.18\scriptsize{±0.7} } & 	 \textbf{71.73\scriptsize{±0.2} } & 	 \textbf{72.9\scriptsize{±0.9} } & 	 {80.78\scriptsize{±0.8} } & 	 {91.03\scriptsize{±0.3} } & 	 {84.73\scriptsize{±0.7} } & \textbf{68.77} \\

VICReg  & {47.91\scriptsize{±1.3} } & 	 \textbf{69.75\scriptsize{±0.2} } & 	 {89.08\scriptsize{±0.1} } & 	  {24.74\scriptsize{±0.5} } & 	  {50.25\scriptsize{±0.8} } & 	  {68.51\scriptsize{±0.6} } & 	  {68.94\scriptsize{±2.9} } & 	  \textbf{82.67\scriptsize{±0.3} } & 	  \textbf{91.21\scriptsize{±0.3} }  & 	  \textbf{85.44\scriptsize{±0.5} } & 67.85 \\

    \end{tabular}
    
}
\caption{
Comparison of our method for different alternate regularization strategies on 4 different datasets.} 
\label{tab:regularizer}
\end{table*}
}

Next, we investigate different regularization strategies. First, we study the choice of applying the contrastive loss on the low-dimensional embedding space rather than the final prediction. In Table \ref{tab:ablation:contrast_aug}, we present the results for this experiment, where we observe a 5.48\% decrease in accuracy, indicating the importance of learning representations by regularizing the embedding space instead of the class predictions. Next, we examine another aspect of our regularization approach by replacing two strong augmentations with one weak and one strong augmentation (similar to FixMatch). This variant results in a 3.78\% drop in accuracy. 

Finally, we show the performance of our method by completely replacing the contrastive loss with other important viable losses, namely, BYOL \cite{byol}, SimSiam \cite{simsiam}, and VICReg \cite{vicreg}. BYOL \cite{byol} learns by predicting the representation of a target encoder (exponential moving average of the online encoder), using the online encoder for different augmentations of the same sample. SimSiam \cite{simsiam} also learns by matching the representations of two augmented samples, but unlike previous methods, it doesn't need negative samples or the target encoder. VICReg \cite{vicreg} learns from unlabelled data by combining three terms: variance, invariance, and covariance regularizations. The results of this experiment are presented in Table \ref{tab:ablation:self-sup}, where
we see that the accuracy of BYOL and SimSiam is considerably lower than the proposed contrastive loss with 46.89\% and 46.50\%. However, we find VICReg to have a competitive performance. To fully understand the performance of VICReg, we present the results for all datasets in Table \ref{tab:regularizer}. This table shows that VICReg achieves better results for a few settings (CIFAR-10 with 250 labelled samples, SVHN with 250 and 1000 labelled samples, and STL-10 with 1000 labelled samples). However, the overall average accuracy with VICReg loss is 0.92\% lower than the proposed contrastive regularizer, and hence, we choose this as the default for the proposed UnMixMatch.

{\renewcommand{\arraystretch}{1.0}   
\begin{table}[t!]
\centering

\setlength{\tabcolsep}{9.5pt}
    \begin{tabular}{l c }
    \textbf{Setting} & \textbf{Accuracy} (\%)  \\
    \shline
    Ours & 47.93\\
    MixUp only & 45.23\\
    w/ Match loss & 44.25\\
    w/ KL-div. loss & 47.02\\
    
\hline
\end{tabular} 
\caption{Study on the relation of our method to prior works.} 
\label{tab:relation_remixmatch}
\end{table}
}

\subsubsection{Rotation Loss.}
We also investigate a few other self-supervised approaches instead of the rotation prediction task, including colorization, jigsaw solving, and image inpainting. The results of this experiment are included in Table \ref{tab:ablation:pre-text}, where rotation prediction shows a clear advantage over other pre-text tasks. Apart from the fact that rotation prediction performs well, it has two additional benefits over other methods. First, it can be integrated into our semi-supervised framework without adding considerable computational overhead. In contrast, tasks like colorization or inpainting require an additional decoder to train, which increases the overall training cost. Second, rotation prediction has already demonstrated effectiveness with other semi-supervised methods such as ReMixMatch on a wide range of datasets.

\subsubsection{Relation to Existing Methods.}\label{sec:ablation-existing}
Our method has a few similarities with some of the previous semi-supervised methods. For example, MixMatch \cite{mixmatch} popularized the use of MixUp \cite{mixup} in the context of SSL. ReMixMatch \cite{remixmatch} built over MixMatch by using MixUp and a rotation prediction loss, making it the most similar to our proposed method. However, our method has certain distinctions from ReMixMatch. Firstly, ReMixMatch only uses MixUp with the supervised module, whereas ours uses RandMixUp. Moreover, our approach uses a contrastive loss as a regularizer, which is applied to the intermediate embeddings, whereas ReMixMatch uses a match loss on the predictions of mixed samples. Finally, ReMixMatch uses KL-divergence loss between the predictions of weakly and strongly augmented samples, which our method does not employ. Overall, recall from Table \ref{tab:ssl_final} that ReMixMatch, on average, shows 5.62\% lower performance than our method. In Table \ref{tab:relation_remixmatch}, we further investigate the differences by incorporating the design choices of ReMixMatch into our method. First, by removing the RandAug from RandMixUp (using MixUp like ReMixMatch), we obtain an accuracy of 45.23\% (2.7\%$\downarrow$). Next, using the match loss of ReMixMatch as a regularizer on the class predictions in our method results in an accuracy of 44.25\% (3.68\%$\downarrow$). Finally, adding the KL-divergence loss to our method further results in a 0.91\% drop in overall performance.

\subsection{Sensitivity Study}\label{sec:sensitivity}
\begin{figure}%
    \centering
    
    \subfloat[Acc vs $\alpha$]{\label{fig:alpha}{\includegraphics[width=.33\linewidth]{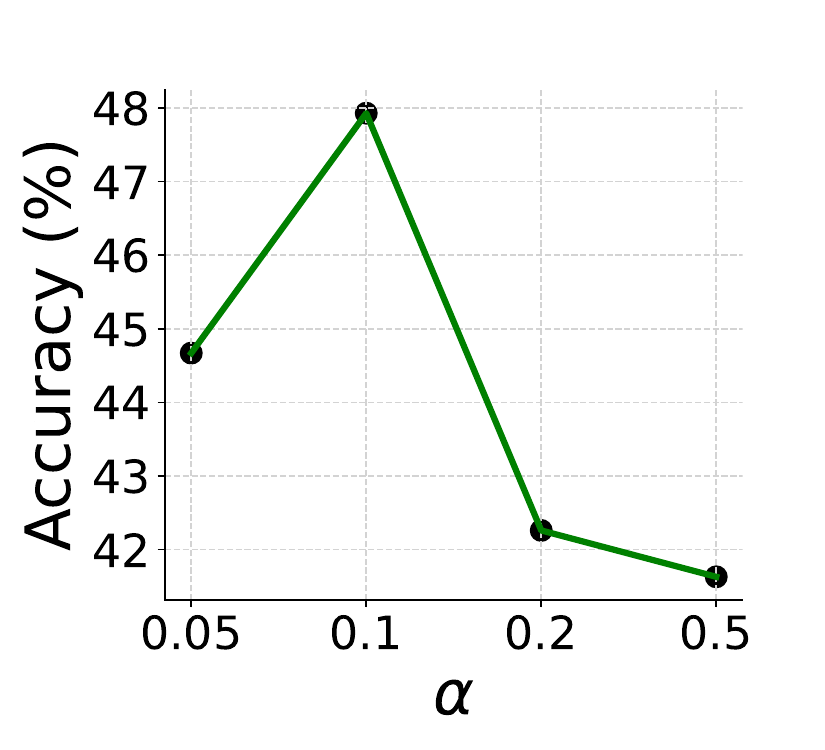} }}%
    \subfloat[\centering Acc vs $\beta$]{\label{fig:beta}{\includegraphics[width=.33\linewidth]{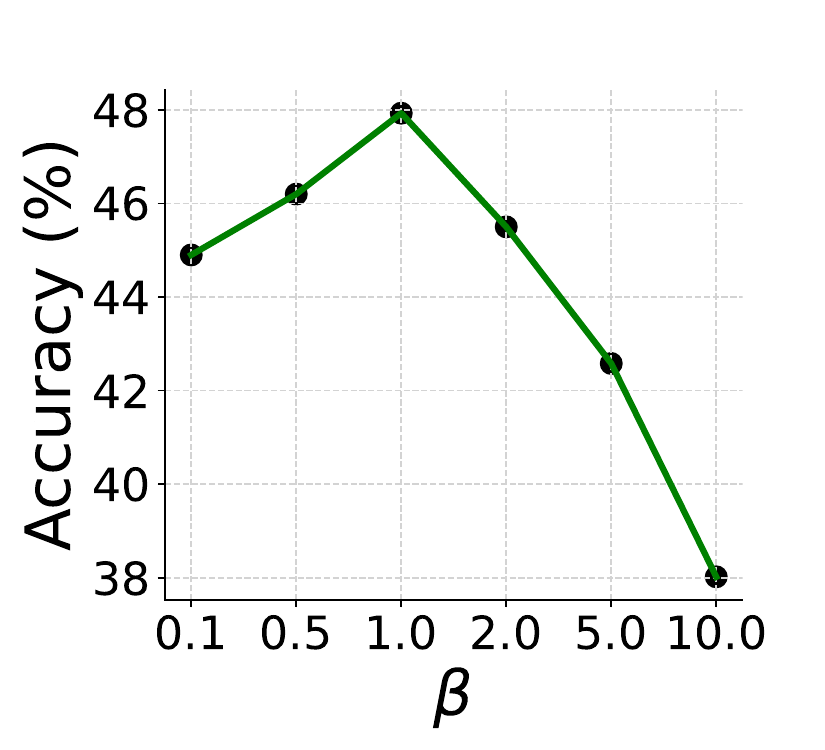} }}%
    \subfloat[\centering Acc vs $\gamma$]{\label{fig:gamma}{ \includegraphics[width=.33\linewidth]{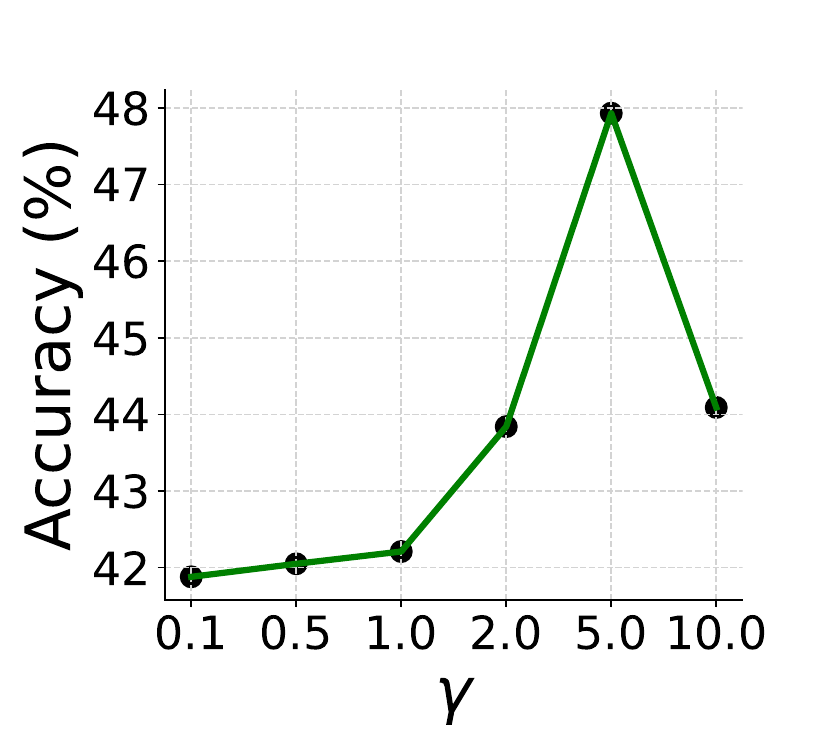} }}%
    \caption{Sensitivity study on important hyper-parameters.}%
    \label{fig:example}%
\end{figure}

Our method involves three important hyper-parameters, namely the $\alpha$ value in MixUp, the contrastive loss weight $\beta$, and the rotation loss weight $\gamma$. Here, we present a sensitivity analysis of different values for these hyper-parameters. As illustrated in Fig. \ref{fig:alpha}, the best accuracy is achieved with an alpha value of 0.1, while very large or small values of $\alpha$ lead to a drop in accuracy. Fig. \ref{fig:beta} indicates that the optimal performance is obtained with a $\beta$ of 1.0. Furthermore, Fig. \ref{fig:gamma} reveals that a higher value of $\gamma$ leads to better performance, highlighting the importance of the self-supervised loss in UnMixMatch.

\section{Conclusion}

    Existing semi-supervised methods struggle to learn when the assumption that the unlabeled data comes from the same distribution as the labelled data, is violated. This work proposes a new semi-supervised method called UnMixMatch for learning from unconstrained unlabelled data. Our method shows large improvements over existing methods and even larger improvements under low-labelled data settings. Our approach also outperforms existing methods on \textit{open set} settings. Most importantly, UnMixMatch scales up in performance when the size of unlabelled data increases. We hope this research will draw attention to this more challenging and realistic SSL setting with unconstrained unlabelled data. 

\noindent \textbf{Limitations.} 
While our comprehensive study on common benchmark datasets demonstrates the effectiveness of UnMixMatch across diverse image domains, some aspects require further investigation for universal applicability. Specifically, the use of hard augmentations and the rotation prediction task might necessitate additional tuning or minor modifications for optimal performance in specialized domains.

\section*{Acknowledgements}
This work was supported by Mitacs, BMO, and Ingenuity Labs Research Institute. We are also thankful to SciNet HPC Consortium for helping with the computation resources.

\bibliography{ref}

\end{document}